\documentclass[conference]{IEEEtran}
\IEEEoverridecommandlockouts

\usepackage{amsmath,amssymb,amsfonts}
\usepackage{graphicx}
\usepackage{booktabs}
\usepackage{array}
\usepackage{xcolor}
\usepackage{url}
\usepackage[colorlinks=true,allcolors=blue]{hyperref}
\usepackage{cite}
\usepackage{bm}
\usepackage{subcaption}
\usepackage{float}  %
\usepackage{verbatim}
\usepackage{tabularx}

\graphicspath{{figures/}}

\newcommand{\R}{\mathbb{R}}

\newcommand{\vcmd}{\bm v_{\mathrm{cmd}}}
\newcommand{\Jhat}{\hat{J}}

\newcommand{\newm}[1]{\mathit{#1}}

\renewcommand{\newm}[1]{#1}

\title{\LARGE \bf Rapid Learning of Dexterous In-Hand Pen Writing through Real-Time Jacobian Estimation}

\author{Kai Stewart$^{1\dagger}$, Yasunori Toshimitsu$^{1\dagger}$, and Robert K. Katzschmann$^{1*}$
\thanks{Supported by the Takenaka Scholarship Foundation, the Max Planck ETH CLS, and the Swiss Government Excellence Scholarship.}
\thanks{$^{1}$All authors are with the Soft Robotics Lab, D-MAVT, ETH Zurich, 8092 Zurich, Switzerland.}
\thanks{$^{\dagger}$These authors contributed equally to this work.}
\thanks{$^{*}$Corresponding author: Robert K. Katzschmann ({\tt\small rkk@ethz.ch}).}
}

\begin{document}
\maketitle

\begin{abstract}
Dexterous in-hand manipulation of a grasped object with an anthropomorphic hand is an unsolved frontier for robot dexterity.
The contact-richness and highly dynamic nature of object-hand interactions tend to require extensive modeling or data-collection efforts for learning-based approaches.
Modern simulators used for reinforcement learning (RL) cannot fully replicate the required contact complexity, while collecting dexterous demonstrations for imitation learning (IL) remains an open problem.
In this research, we present an embodied control approach based on real-time task Jacobian estimation of the combined hand and object system on the physical robot. Using only the CPU on a laptop, the proposed controller begins in-hand pen writing after approximately 18\,s of initialization and continues to adapt online, without an analytic hand--object kinematic/contact model, simulation training, or precollected task demonstrations.
We demonstrate that the same estimator/controller formulation works on three anthropomorphic robotic hand systems (one physical, two simulated) to show human-like, in-hand articulation of a grasped pen by an embodiment-independent formulation.
Sub-millimeter in-plane precision (mean 0.6 mm across runs) is achieved across letters and shapes written in the air and on paper on a physical robot.
To our knowledge, this is the first demonstration of an anthropomorphic hand writing arbitrary single-stroke trajectories with a grasped pen through purely in-hand motion, and it showcases an alternative to compute- and data-heavy approaches such as RL and IL for achieving dexterous manipulation through computationally simple and data-efficient algorithms. 
\footnote{\url{https://srl-ethz.github.io/rapid-dexterous-writing/}}
\end{abstract}

\bstctlcite{BSTcontrol}

\section{Introduction}
\label{sec:intro}

As robots evolve from specialized machines in controlled environments into
generalist machines for workplaces and homes, learning-based methods such as reinforcement learning (RL) and imitation learning (IL) have become prominent approaches for manipulation~\cite{handa2023dextreme, ilsurvey2025}. As workplaces and homes are built around the human hand, acting in them
entails manipulating everyday objects with comparable dexterity. However, contact-rich in-hand
manipulation --- repositioning or articulating a grasped object with the fingers alone ---
remains an active area of research.  

\begin{figure}[ht]
  \centering
  \includegraphics[width=\linewidth]{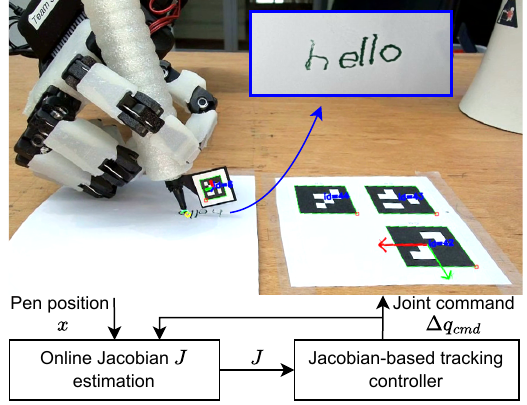}
  \caption{The biomimetic robotic hand writing text with the proposed method. Holding a real pen in a human-like grasp, the hand traces letters on paper by finger motion alone (no
  wrist or arm), driven by a command-space task Jacobian estimated online from vision,
  without the hand's kinematic/contact model, training in simulation, or precollected task demonstrations. The hand was mounted on a robot arm used only to move the hand to the next letter, providing no assistance during the writing motion.}
  \label{fig:teaser}
\end{figure}

RL can produce striking in-hand results
in controlled scenarios, but relies on heavy domain randomization to close the residual sim-to-real gap due to the dynamic nature of the contacts~\cite{handa2023dextreme}. IL needs no
contact model but is bottlenecked by demonstration data~\cite{ilsurvey2025}, and
in-hand manipulation is where it frequently struggles: with the fingers wrapped around the
object, hand and object occlude one another, so no purely visual recording, not
even the egocentric human video increasingly used to supplement robot
data \cite{Punamiya2026-ti}, sees every finger and contact a demonstration must
capture. The human motion data must also be retargeted to robot hands with different kinematics, which remains challenging~\cite{pan2025spider}. Both approaches have shown impressive
results, but remain far from solving dexterous manipulation and demand massive
modeling/simulation or data-collection effort. One of the reasons for this is the inherent complexity arising from the constantly changing contact dynamics, which make the map from finger actuation to object motion nonlinear, configuration- and time-dependent \cite{ilsurvey2025}.

Here we present an online task-Jacobian estimation approach to in-hand pen writing. Rather than analytically modeling the
contacts or collecting demonstrations, we extend previous approaches which estimate the manipulator Jacobian for visual servoing of robot arms \cite{hosoda1994, dije2022} to an in-hand manipulation task. By estimating the local differential
relationship between actuation and object motion directly on the robot, we achieve online closed-loop control without an analytic hand--object kinematic/contact model, simulation training, or precollected task demonstrations

We demonstrate this on the $17$-DoF, tendon-driven, anthropomorphic ORCA
hand~\cite{orcahand2025}: grasping a pen in a compliant 3D-printed casing, it writes by finger motion alone (Fig.~\ref{fig:teaser}), tracing closed shapes and letter glyphs. We chose robotic writing as
an exacting test of in-hand control: the hand must hold onto the pen, forming multiple closed kinematic chains in a redundant structure, which makes direct modeling difficult. The stroke is visible, so success and
failure are legible and directly measurable, and doing it on paper rather than in
the air shows its practical usefulness. The pipeline is deliberately
light, needing only the CPU of a laptop and a simple webcam, yet tracks the 
path with sub-millimeter in-plane error.

\noindent{Contributions:}
\begin{itemize}
  \item \emph{An anthropomorphic hand writing arbitrary single-stroke trajectories on paper through in-hand articulation of a real pen}, via a lightweight command-space task Jacobian estimated and updated through real-time interaction, without an analytic hand--object model, simulation training, or precollected task demonstrations. The same estimator/controller formulation is demonstrated on three robot platforms (one physical, two in simulation), and released 
  \href{https://github.com/srl-ethz/dexterity_from_jacobian}{as open-source code}.
  \item \emph{A Jacobian-estimation control architecture for redundant in-hand systems}: a short grip-pose excitation bootstraps the map, sustained learning keeps it accurate, and a damped task-space controller commands the joints while its nullspace maintains a stable grip (Sec.~\ref{sec:rls}--\ref{sec:control}).  
  \item \emph{An evaluation that characterizes the method's limits}: ablation tests that compare the in-plane tracking metrics and robustness, revealing the elements that are essential to achieve robotic writing.
\end{itemize}

This work shows that online, real-time learning is a viable route to in-hand manipulation, especially where a lightweight, adaptive controller is needed.

\section{Related Work}
\label{sec:related}

\subsection{Online Jacobian estimation and visual servoing}
Uncalibrated visual servoing learns the differential map between sensed features
and actuation online, with no camera or kinematic
calibration. Hosoda and
Asada~\cite{hosoda1994} introduced the canonical recursive-least-squares (RLS)
estimator with a forgetting factor, the direct ancestor of our update.
Qian and Su~\cite{qiansu2002} estimated the image Jacobian online with a Kalman
filter.

Recent attempts have extended this to a more generalized descriptor of the Jacobian, obtaining a dense, per-pixel image Jacobian \cite{dije2022} or a visuomotor Jacobian field defined on all 3D points along the robot \cite{li2025}.

However, the visuomotor Jacobian field required extensive data to compute, around 2--3 hours of robot motion from a multi-view RGBD camera setup~\cite{li2025}.
All of these past methods showed no or only limited application to tool manipulation: Toshimitsu et al. showed bimanual broom tip manipulation \cite{dije2022} and Li et al. applied the Jacobian-field-based control to grasp a tool and push an apple along a table \cite{li2025}, in one of their demonstrations.

Prior work has also estimated the inverse of the Jacobian instead of the Jacobian itself, with little \cite{grace2024} or no \cite{grace2025} a priori information, and applied it to a writing task in air. By directly estimating the inverse Jacobian, a direct map from task space to command (joint) space can be obtained, eliminating the need to computationally invert the Jacobian matrix. However, as only its inverse is estimated, it is not clear how to compute and utilize the nullspace of the Jacobian, limiting the extendability to redundant structures.

\subsection{In-hand writing as a benchmark}
Handwriting and pen-tip path tracing are recurring benchmarks for within-hand manipulation because they require coordinated finger motion while maintaining a stable grasp.
We focus on systems that use primarily the fingers to articulate the pen, distinguishing them from those that use the arm to generate the writing motion.
Several real-hardware studies manipulate a grasped pen without direct Cartesian pen-tip feedback, instead relying on open-loop pose transitions \cite{Zhou2019-ei} or control in task-specific, tendon-derived synergy spaces \cite{Malhotra2012-rh}. Other studies evaluate free-space path tracing, moving a grasped pen or block through the air rather than depositing ink on a writing surface \cite{hu2023slender,morgan2019,grace2024,grace2025}.

The physical writing demonstrations by Rombokas et al. and Zhao et al. are the closest to our setup, using respectively a biomimetic musculoskeletal hand \cite{Rombokas2011-qt} and a three-fingered gripper \cite{Zhao2025-ar} to grasp a pen to write on a surface.
Rombokas et al. use a synergy-based controller to learn the mapping between synergies and task space on the real robot, with the best plotted trial achieving under 0.2 mm error, all evaluated on the letter 'S'. Their evaluation was run only on a single letter and not arbitrary trajectories, and their data collection procedure for the recording duration or exact demonstration motions is not specified, limiting exact reproduction on a new hand~\cite{Rombokas2011-qt}.
Zhao et al. transfer an RL policy trained in simulation to a real robot with tactile sensors. The physical system writes using several tools and surfaces, but pen-tip tracking errors are reported only for a simulated robot and no path-tracking errors are reported for real-hardware evaluation. The authors also note that complex trajectories requiring high precision remain difficult to execute consistently \cite{Zhao2025-ar}.

We show broad adaptability of our method: after ${\sim}18$\,s of initialization, the physical Orca hand writes the whole alphabet while continuously updating its Jacobian estimate. We also demonstrate the same estimator/controller formulation on two other hands in simulation, with the code released as an open-source implementation.

\section{System Architecture}
\label{sec:system}
\subsection{ORCA hand}
We use the ORCA hand~\cite{orcahand2025}, an anthropomorphic,
tendon-driven hand with 17 actuated degrees of freedom. The pen is held in a fixed
power-precision grip; the wrist, ring, and pinky (7 joints) are held fixed, as they are not relevant
to pure in-hand pen writing, while the thumb, index, and middle fingers, the $N=10$ joints
we control (Sec.~\ref{sec:problem}), articulate the pen. The pen is wrapped
in a custom soft TPU sleeve that enlarges its effective diameter to roughly $4\times$
that of the bare pen, giving the hand a larger, more secure grip surface and adding
mechanical compliance at the pen--grip interface.

\subsection{Perception for pen-tip tracking in the paper frame}
\label{sec:perception}
A webcam tracks an ArUco marker attached to the pen and a board of three markers taped to the desk, defining the paper frame. The pen-tip position is recovered using a fixed marker-to-tip offset and transformed into the paper frame, which is used for all downstream estimation and control.
Fig.~\ref{fig:teaser} shows the view from the webcam with the marker detection and pen tip position recovery results overlaid.

The resulting pen-tip trajectory is smoothed using a constant-velocity Kalman filter with innovation-based outlier rejection. Camera timestamps are retained throughout the pipeline to provide a common time axis for estimation and control.

\section{Method}
\label{sec:method}

\begin{figure}[t]
  \centering
    \includegraphics[height=0.5\textheight]{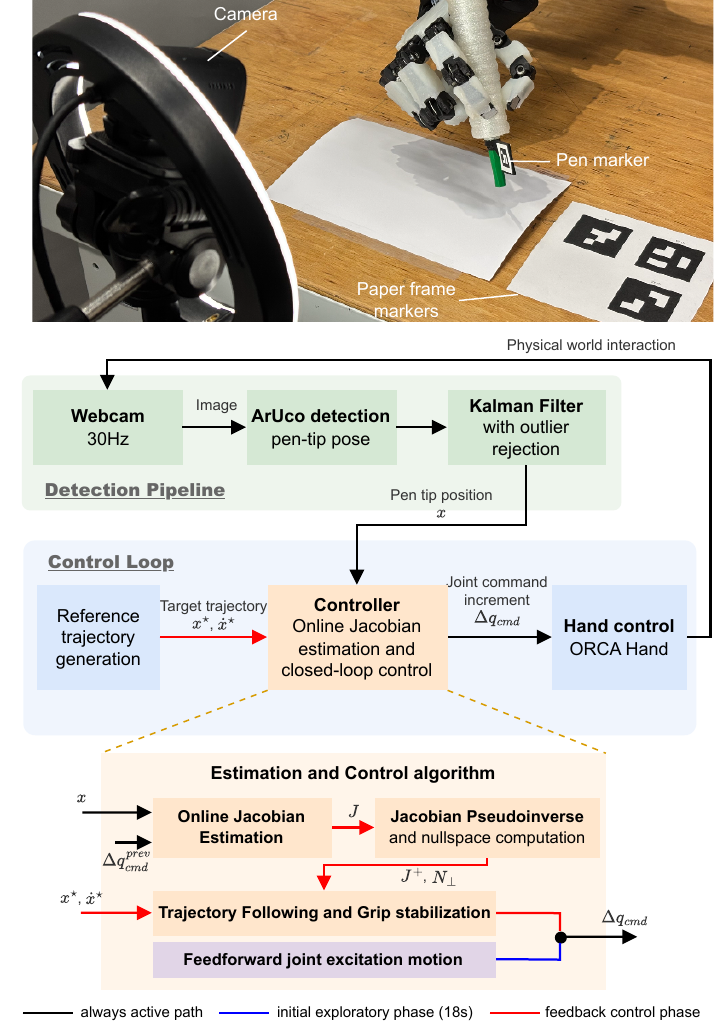}
  \caption{System Architecture. A perceive--estimate--act loop tracks the pen tip,
  updates the task Jacobian $J$ online, inverts it and computes its nullspace. These define the desired motor commands, which are then sent to the ORCA hand. A separate controller initially moves the fingers in a predetermined trajectory to bootstrap the Jacobian estimation.}
  \label{fig:pipeline}
\end{figure}

We transfer the classical online forward-Jacobian estimator from serial-arm visual servoing to a harder setting, a redundant, tendon-driven in-hand system, and make it work there through two additions: a deliberate excitation phase and nullspace grip stabilization.

Fig.~\ref{fig:pipeline} gives a high-level overview of our control pipeline: the pen tip position is computed from the camera image and passed downstream, where the  command-space task Jacobian is estimated, which the controller combines with the reference trajectory to define the joint commands. The hardware then executes the received joint commands, and closes the feedback loop by feeding the current state back. The pipeline is event-driven (one control
step per pen-tip update), so the control rate is perception-limited at a consistent
${\approx}15$\,Hz, below the camera's nominal $30$\,Hz.

\subsection{Problem formulation}
\label{sec:problem}

We control the planar position of the pen tip on the paper. Let
$\bm q\in\R^{N}$ be the vector of actuated finger-joint positions the controller
commands, and $\bm x\in\R^{m}$ the pen-tip position in the paper plane. In our
default configuration $m=2$ (the in-plane coordinates $x,y$) and $N=10$ finger
joints: the four thumb joints and three joints each of the index and middle
fingers; the wrist, ring, and pinky are held fixed.
While writing each letter, we use neither the wrist nor the arm, in order to study \emph{pure} in-hand manipulation, in which the pen is articulated solely by the
fingers within a human-like grasp. Under this self-imposed limitation, and given the bounded dexterity of the utilized robotic hardware, we do not control the $z$-height of the pen tip: we treat it as uncontrolled drift, exclude it from our in-plane error metric (Sec.~\ref{sec:results}), and report its observed magnitude separately.

For on-paper runs, the sheet is folded over itself or taped on the table such that the middle area of the paper bulges upwards by about a centimeter, absorbing the uncontrolled ${\sim}2$--$3$\,mm $z$ drift so that the pen tip remains in contact with the paper.

Locally, joint and task velocities are related by the task Jacobian
$J\in\R^{m\times N}$,
\begin{equation}
  \dot{\bm x} \;=\; J\,\dot{\bm q},
  \label{eq:jacobian}
\end{equation}
which for an in-hand grasp is unknown and configuration-dependent. Our goal is to (i) estimate $J$ online from
the motion command $\Delta q_{\mathrm{cmd}}$ and observed pen motion $\dot{\bm x}$, with no analytic kinematic model of the
tendon-driven hand, and (ii) invert the estimate, with light damping for
robustness while it converges, to track a desired writing trajectory $\bm x^{\star}(t)$.

\subsection{Recursive online task Jacobian estimation}
\label{sec:rls}

The online task Jacobian $J$ estimation follows the formulation by Toshimitsu et al.~\cite{dije2022} (in turn based on Hosoda and
Asada~\cite{hosoda1994}). We formulate the command-space task Jacobian estimation as a recursive least-squares (RLS) / Kalman-filter-style problem. We approximate the covariance matrix $P$ by a diagonal matrix $P=\text{diag}(\bm p)$ and therefore store only its diagonal entries $\bm p$. The values are initialized as $\bm p = p_{\mathrm{init}}\bm 1$.

Updates run only when at least one joint is actively
moving ($\max_i|\dot q_i|$ above a threshold), preventing the estimate from drifting
on observation noise when the hand is nearly still. This gate is evaluated
on the measured joint velocities, whereas the regressor below uses the commanded
increment.

The joint velocity $\bm{\dot q}$ used in the command-space task Jacobian estimation is computed from the incremental joint command of the previous step, $\Delta \bm q_{\mathrm{cmd}}^{\mathrm{prev}} / \Delta t$, rather than from measured joint velocities: since the controller inverts the estimate to produce commands, the command-to-task-motion mapping is the operationally relevant one, and the estimate absorbs unmodeled actuator and transmission effects. Measured joint velocities are also substantially noisier on our platform, and using them as the regressor destabilized the estimate in preliminary experiments.
At each step, the covariance is first inflated by a forgetting factor
$\lambda\in(0,1]$ so the estimate does not fully converge and can track slowly changing dynamics,

\begin{equation}
    \bm p \leftarrow \frac {\bm p} \lambda,
\end{equation}
then the Jacobian and covariance are updated from the new $(\dot{\bm q},\dot{\bm x})$ pair,
\begin{equation}
    J \leftarrow J + \frac{(\dot{\bm x} - J \bm{\dot q})\,(\bm p \odot \bm{\dot q})^{\!\top}}{\bm p^{\top}(\bm{\dot q}\odot\bm{\dot q}) + r},
\end{equation}
\begin{equation}
\begin{split}
    \bm p &\leftarrow \bm p \odot \Bigl(\bm 1 - \frac{\bm p\odot\bm{\dot q}\odot\bm{\dot q}}{\bm p^{\top}(\bm{\dot q}\odot\bm{\dot q}) + r}\Bigr),\\
    \bm p &\leftarrow \mathrm{clip}(\bm p, p_{\mathrm{floor}}, p_{\mathrm{init}}).
\end{split}
\end{equation}

Here $r$ is the observation-noise variance and $\odot$ the element-wise product.
Clipping $\bm p$ keeps the updates stable while preserving adaptivity. Parameter
values are in Table~\ref{tab:params}.

\subsection{Excitation phase}
\label{sec:excite}
Because the fingers and grasped object form multiple closed kinematic chains, arbitrary joint excitation need not generate informative task-space motion and can instead perturb the grasp.
We therefore begin each run with a short \emph{excitation} phase
(${\sim}18$\,s; Table~\ref{tab:params}) sending a predefined trajectory to the joints, bootstrapping the Jacobian estimate before
tracking. The Jacobian is initialized with zeros, and the robot hand commands are overridden while the Jacobian estimation process is kept running, and the hand is swept through six
grip poses along a Catmull--Rom spline~\cite{catmullrom1974} (${\sim}15$\,s, then
${\sim}3$\,s to settle), injecting rich, $C^1$-continuous joint motion so the
estimator observes a well-conditioned set of $(\dot{\bm q},\dot{\bm x})$ pairs.
We found
this spline interpolation preferable to piecewise-linear interpolation (velocity discontinuities at
waypoints) or smoothstep (velocity vanishing at every waypoint), which excite the
estimator less effectively. 
The six poses are fixed joint-space waypoints, obtained once by manually
posing the gripping hand and recording the joint values. The poses were chosen so that,
while keeping the grasp secure, the pen tip spans the reachable task space and in
particular both commanded in-plane directions. This manual one-time setup took only minutes
and involved no systematic optimization, and the same waypoints are reused unchanged in
all physical ORCA experiments.

\begin{table}[t]
\centering
\caption{Default parameters for the physical ORCA experiments, unless otherwise noted.}
\label{tab:params}
\renewcommand{\arraystretch}{1.1}
\setlength{\tabcolsep}{4pt}
\footnotesize
\begin{tabular}{@{}l l p{1.3in}@{}}
\toprule
Symbol / name & Value & Description \\
\midrule

\multicolumn{3}{@{}l}{\textit{Jacobian estimator}}\\
$N$               & $10$                   & actuated finger joints \\
$m$               & $2$                    & task dim.\ (in-plane) \\
$r$               & $4.0$                  & obs.\ noise scalar \\
$p_{\mathrm{init}}$ & $0.1$                & init.\ covariance \\
$p_{\mathrm{floor}}$ & $0.01$              & covariance floor \\
$\lambda$         & $0.999$                & forgetting factor \\
\midrule
\multicolumn{3}{@{}l}{\textit{Controller \& path}}\\
$\varepsilon$     & $0.003$                & pseudo-inverse damping \\
$k_{\mathrm{pb}}$ & $0.05$                 & posture pullback gain \\
$K_p,K_i,K_d$     & $100,\,2,\,6$          & path PID gains \\
excitation time   & $18$\,s                & $15$\,s sweep ${+}\,3$\,s settle \\
$\lVert \dot{\bm x}^\star \rVert_2$ & $8\times10^{-4}$\,m/s  & writing speed \\
loop rate         & ${\sim}15$\,Hz         & perception-limited (camera nominal $30$\,Hz) \\
\bottomrule
\end{tabular}
\end{table}

\subsection{Writing-path generation}
\label{sec:path}
After the excitation phase the tracking phase begins, in which the writing-path generator sends the task command to the controller in each step.
The trajectories are each reduced to a planar curve in the paper frame and traversed at constant tangential speed (so pen-tip speed is independent of curvature).
We define the letters \textsc{A--Z} as cubic-B\'ezier glyphs, while also supporting arbitrary single-stroke shapes within the writable range of the pen, such as parametric shapes (circle, flower, heart, star) and free-form \textsc{svg} drawings. 

The task-space command sent to the controller is a PID-plus-feedforward law on the tracking error
$\bm e=\bm x^{\star}-\bm x$,

\begin{equation}
\begin{aligned}
  \vcmd = {}& K_p\,\bm e + K_i\,\mathrm{clip}\!\Bigl(\textstyle\int \bm e\,dt,\ \pm c\Bigr)\\
        &{}+ K_d\,\tfrac{\bm e-\bm e_{\mathrm{prev}}}{\Delta t} + \bm \dot x^\star,
\end{aligned}
  \label{eq:pid}
\end{equation}
with anti-windup clamp $c$ on the integral and feedforward $\bm \dot x^\star$ equal to
the path tangent velocity at the current target.

\subsection{Control law}
\label{sec:control}

On each task command we first update the Jacobian (Sec.~\ref{sec:rls}), then invert
the current estimate with a damped (Tikhonov) right pseudo-inverse $J^{+}$, adding a posture stabilization
term projected through an approximte nullspace operator $N_{\!\perp}$:

\begin{align}
  J^{+} &= J^{\top}\bigl(JJ^{\top}+\varepsilon I_m\bigr)^{-1},
  \qquad N_{\!\perp}=I_N-J^{+}J, \\
  \Delta \bm q &= \Bigl(\underbrace{J^{+}\vcmd}_{\text{task}}
            + \underbrace{k_{\mathrm{pb}}\,N_{\!\perp}\,(\bm q_0-\bm q)}_{\text{posture stabil.}}
            \Bigr)\,\Delta t .
  \label{eq:control}
\end{align}

The damping $\varepsilon$ regularizes the inverse, bounding the command when $J$ is
poorly conditioned. The posture term pulls the hand back toward its initial grip pose $\bm q_0$ through the approximate nullspace of $J$, so that the posture can be adjusted while minimizing interference with the task.

The task command $\vcmd$ comes from the writing-path generator (Sec.~\ref{sec:path}), and parameter values are in Table~\ref{tab:params}.

Finally $\Delta \bm q$ is clipped to a per-joint velocity limit and low-pass filtered
\begin{equation}
  \Delta \bm q_{\mathrm{cmd}} = 0.8\,\Delta \bm q + 0.2\,\Delta \bm q_{\mathrm{cmd,prev}},
\end{equation}
before being integrated into the absolute joint-position command sent to the
hand.

\section{Results}
\label{sec:results}

Each run establishes the grip, runs the excitation phase
(Sec.~\ref{sec:excite}), then tracks a reference path in air or on paper: a
closed shape, a letter, or a free-form SVG drawing
(Sec.~\ref{sec:path}). Ablations (Sec.~\ref{sec:ablations}) change one
component per run against the fully configured system.

We report in-plane tracking error $\lVert \bm x^\star - \bm x\rVert_{xy}$
(mean and 95th percentile, post-excitation, with per-axis $x,y$ deviations), in
air and on paper; robustness (whether a run completes or diverges, and whether
accuracy stays stable over long runs) and the $z$ drift.

\begin{figure}[t]
    \centering
    \begin{subfigure}[b]{0.32\linewidth}
        \centering\includegraphics[width=\linewidth]{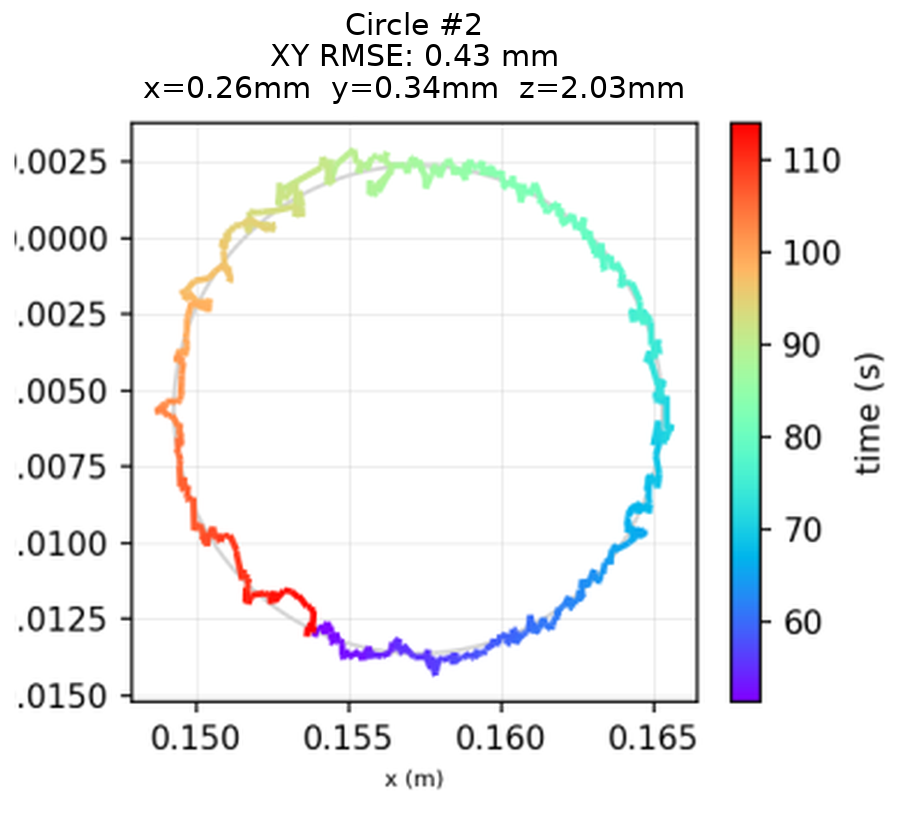}
        \caption{Circle, air.}\label{fig:cleancircles}
    \end{subfigure}\hspace{0.02\linewidth}
    \begin{subfigure}[b]{0.32\linewidth}
        \centering\includegraphics[width=\linewidth]{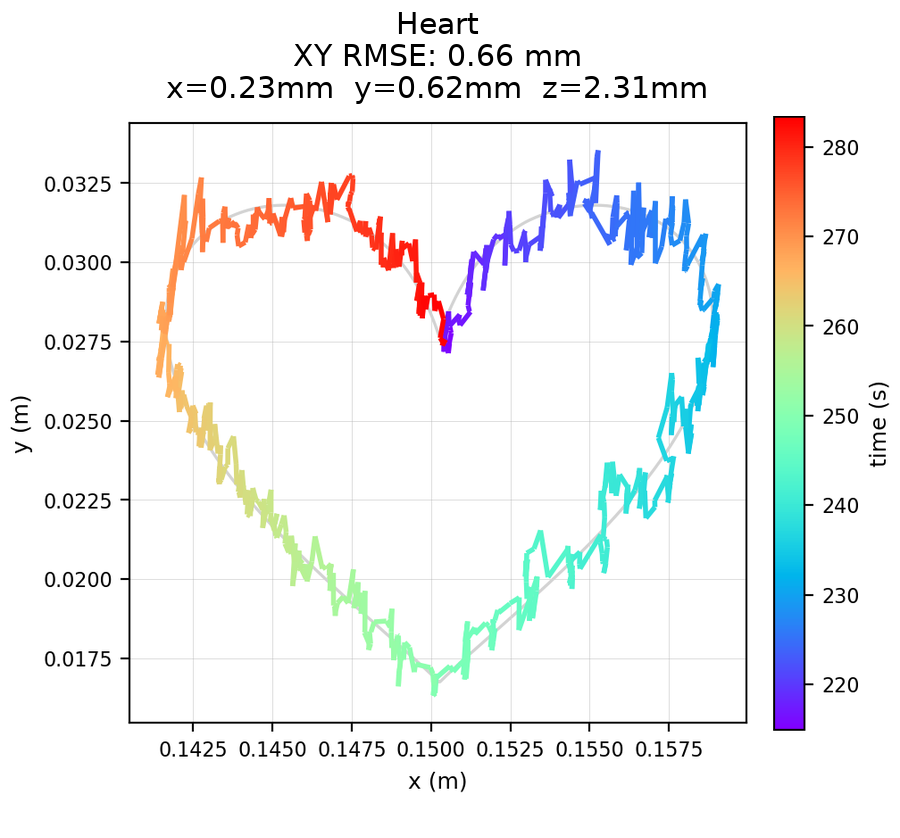}
        \caption{Heart, \textsc{svg}.}\label{fig:heart-svg}
    \end{subfigure}
    \caption{(a)~A circle traced in air and (b)~a free-form closed shape
    sampled from an \textsc{svg} outline. In-plane tracking stays sub-millimeter throughout these runs. }
    \label{fig:trajectory}
\end{figure}

\begin{figure}[t]
  \centering
  \includegraphics[width=\linewidth]{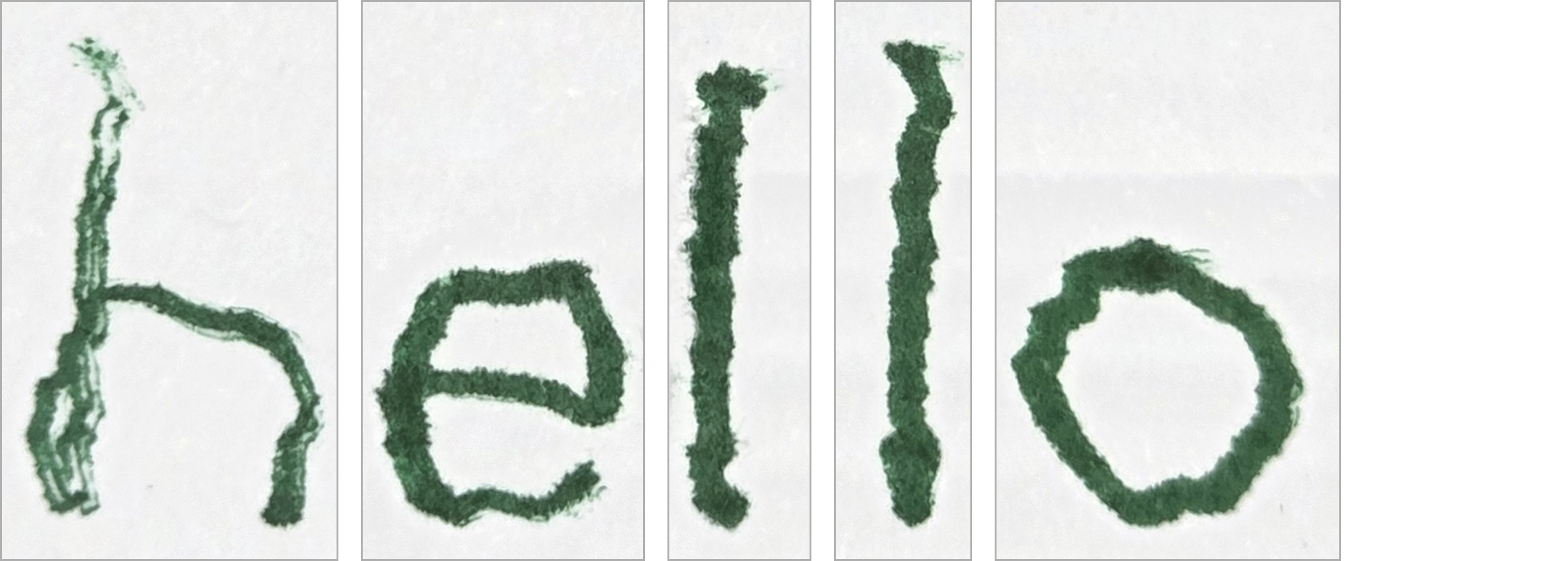}
  \includegraphics[width=\linewidth]{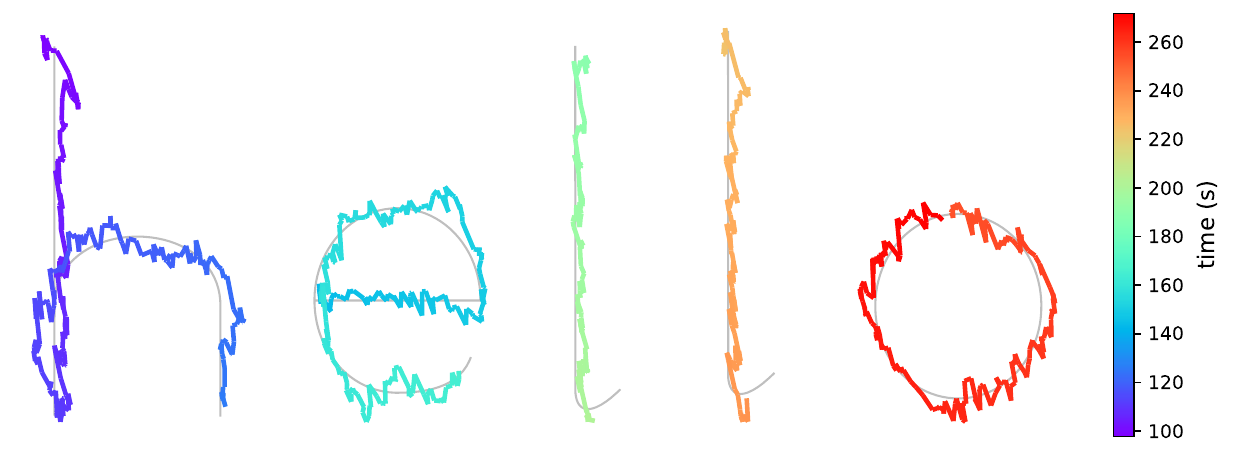}
  \caption{On-paper writing. \emph{Top:} ink deposited by the pen
(letters cropped and re-spaced to match the panels below; all from one run);
\emph{bottom:} the corresponding pen-tip tracking for the same run. The hand writes ``hello'' letter by letter by
finger motion alone. In-plane tracking stays consistently sub-millimeter (per-letter
XY RMSE $0.55$--$0.75$\,mm), comparable to the in-air metric
(Table~\ref{tab:accuracy}). The compliant paper mount absorbs the out-of-plane
drift.}
  \label{fig:onpaperfig}\label{fig:onpaper}
\end{figure}

\begin{figure*}[t]
  \centering
  \includegraphics[width=\textwidth]{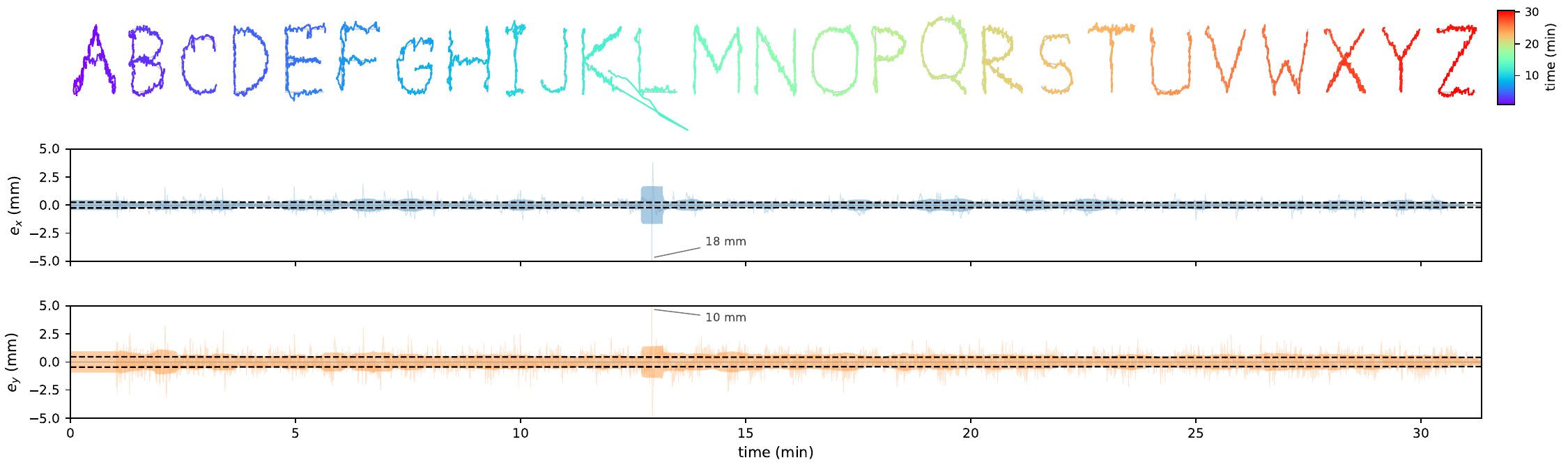}
  \caption{Long-horizon repertoire and in-plane robustness, one continuous run, writing in the air.
  \emph{Top:} all $26$ letters, each a single continuous arc-length B\'ezier glyph
  traced by in-hand finger motion (commanded target in gray; achieved pen tip colored
  by time). \emph{Bottom:} in-plane
  tracking error per axis (faint signed trace; filled
  $\pm$RMS envelope over a ${\sim}30$\,s window; dashed robust Theil--Sen trend). The mean error for each letter stays sub-millimeter apart from a single transient at "K" and its envelope
  narrows over the ${\sim}31$\,min of writing rather than drifting up. The spike at ``K'' results from a sudden finger motion that
  jerks the pen off the path (peak in-plane norm ${\sim}20$\,mm, above $5$\,mm for
  ${\sim}1$\,s) and the controller recovers on its own within the same letter.}
  \label{fig:repertoire}%
  \label{fig:alphabet}%
  \label{fig:durability}%
\end{figure*}

An online forward Jacobian estimated from observed joint and pen-tip motion was enough to drive contact-rich in-hand
writing on the anthropomorphic ORCA hand.
By finger motion alone, the hand traces closed shapes and SVG paths (Fig.~\ref{fig:trajectory}), writes on paper (Fig.~\ref{fig:onpaper}), ran a continuous pass through all $26$
letters of the Latin alphabet (Fig.~\ref{fig:alphabet}).

\subsection{Tracking accuracy}
Within the controlled 2-D action space the writing is precise. Across the
twenty-two post-excitation in air runs, the in-plane error averages
${\sim}0.62\pm0.12$\,mm (range ${\sim}0.39$--$0.86$\,mm), with 95th percentile
${\sim}1.4$\,mm and per-axis std sub-millimeter; the best run reaches
${\sim}0.39$\,mm (Table~\ref{tab:accuracy}).

\emph{On paper}, in-plane accuracy is unchanged. Across sixteen on-paper
runs (eight single shapes and letters, and eight writing ``hello'' letter by
letter, a pre-programmed arm motion shifting the hand between letters while the
fingers alone produced every stroke and the Jacobian estimate persisted across
the repositioning), the error averages ${\sim}0.67\pm0.08$\,mm (range
${\sim}0.57$--$0.82$\,mm, 95th percentile ${\sim}1.4$\,mm), the folded paper
absorbing the uncontrolled out-of-plane drift, so writing with real surface
contact does not meaningfully degrade in-plane tracking.

Pooled over all
$n{=}38$ fully configured runs, in air and on paper, the error averages
${\sim}0.64\pm0.10$\,mm.

\begin{table}[h]
\centering
\caption{In-plane ($x,y$) pen-tip tracking error, in air and on paper.}
\label{tab:accuracy}
\renewcommand{\arraystretch}{1.05}
\setlength{\tabcolsep}{5pt}
\footnotesize
\begin{tabular}{@{}lcc@{}}
\toprule
Metric & air ($n{=}\newm{22}$) & paper ($n{=}\newm{16}$) \\
\midrule
Mean error                   & ${\sim}\newm{0.62\pm0.12}$ & ${\sim}\newm{0.67\pm0.08}$ \\
\quad(run-to-run range)      & $\newm{0.39}$--$\newm{0.86}$ & $\newm{0.57}$--$0.82$ \\
Mean error (best run)        & ${\sim}\newm{0.39}$ & ${\sim}\newm{0.57}$ \\
95th percentile              & ${\sim}1.4$  & ${\sim}\newm{1.4}$ \\
Per-axis std, $x$            & ${\sim}\newm{0.34}$ & ${\sim}\newm{0.42}$ \\
Per-axis std, $y$            & ${\sim}\newm{0.7}$ & ${\sim}\newm{0.68}$ \\
Out-of-plane $z$ drift       & ${\sim}\newm{0.6}$--$\newm{3}$ & ${\sim}\newm{2}$--$\newm{3}^\ddagger$ \\
\bottomrule
\end{tabular}
\\[2pt]
{\scriptsize All entries in mm, across $n{=}38$ post-excitation runs. The out-of-plane $z$ axis is outside the 2-D action space and excluded
from the tracking metric. $^\ddagger$Excludes the eight arm-repositioning runs, whose $z$ spans
the lift height between letters.}
\end{table}

\subsection{Ablations}
\label{sec:ablations}
The writing result rests on two controller choices (a \emph{continuously}
updated Jacobian and a nullspace grip regularization), and removing either
degrades performance (Table~\ref{tab:ablation}). All conditions share the
platform, grip, and task set of the full configuration; one component is
changed at a time.

\begin{figure}[htb]
  \centering
  \includegraphics[width=\linewidth]{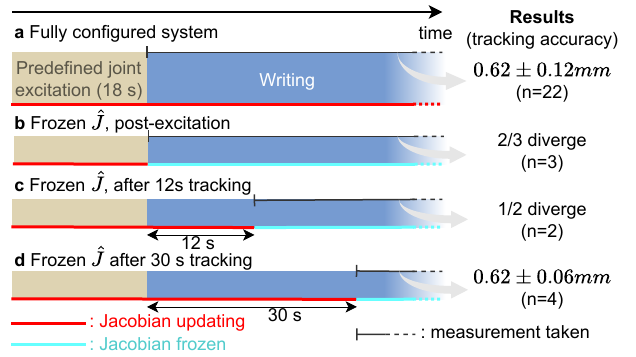}
  \caption{Visual overview of Jacobian freezing conditions for the ablation tests, whose results are in Table~\ref{tab:ablation}.}
  \label{fig:experiment_ablations}
\end{figure}

\emph{Online vs.\ frozen Jacobian.} This is the central ablation, and it
separates three regimes. The experiment conditions are visually summarized in Fig.~\ref{fig:experiment_ablations}. \emph{(i)} An estimate frozen at its post-excitation
value, with RLS disabled, fails: two of three runs left the path within two
minutes (one within ${\sim}40$\,s), the third surviving only at an elevated
${\sim}1.0$\,mm. \emph{(ii)} Freezing after just ${\sim}12$\,s of writing was little
better: one run diverged, the other degraded to ${\sim}1.1$\,mm. The excitation
phase alone does not yield a usable map. A modest amount of
\emph{on-the-job} learning, however, can suffice: \emph{(iii)} frozen after ${\sim}30$\,s of
tracking the estimate held the continuous baseline ($0.62\pm0.06$\,mm, $4/4$
completing), and generalized across trajectories: switching the frozen map from
the circle it had learned to unseen letters left accuracy unchanged
($0.54$--$0.64$\,mm). Even so, continuous updating is best suited in that it is the only mode that can \emph{adapt} to disturbances the frozen
map never saw (e.g. grip slippage as in Fig.~\ref{fig:alphabet}). Continuous, on-the-job adaptation is what makes the online estimate work and keeps it robust.

\emph{Grip regularization.} With the nullspace posture term disabled, the
grasp destabilizes and the fingers lose the pen: across ten such
runs, four failed outright (two catastrophically, at mean
${>}12$\,mm with p95 up to ${\sim}140$\,mm; two aborted within a minute) and
two more degraded severely (means $2.1$ and $3.3$\,mm), while only four completed near
baseline ($0.55$--$0.87$\,mm). The grip regularization is thus what keeps the
grasp stable enough to write reliably.

\emph{Writing speed.} Writing is deliberately slow
($\lVert \dot{\bm x}^\star \rVert_2=8\times10^{-4}$\,m/s). Raising the commanded speed nonlinearly trades
accuracy for throughput: at
$2\times$, one run kept its grip and tracked at ${\sim}0.8$\,mm while its
repeat at the same setting degraded to a $2.3$\,mm mean (p95 $6.6$\,mm); at
$3\times$ the run degraded to ${\sim}2.1$\,mm, yet a $4\times$ run still held
${\sim}1.1$\,mm --- while runs that ramped the speed progressively mid-run (the speed was increased $2\times$, $3\times$, $4\times$, ... sequentially during writing) all had large transients or
diverged (excursions to ${\sim}57$\,mm). 

\begin{table}[htb]
\centering
\caption{Component ablations (in-plane $x,y$ tracking error). Compared against
the twenty-two fully configured runs; one component is changed per condition.
``Diverges'' means the pen tip left the path and did not recover, so a mean is
not meaningful.}
\label{tab:ablation}
\renewcommand{\arraystretch}{1.15}
\setlength{\tabcolsep}{4pt}
\footnotesize
\begin{tabularx}{\columnwidth}{@{}>{\raggedright\arraybackslash}Xcc>{\raggedright\arraybackslash}X@{}}
\toprule
Condition & Mean $\pm$ std (mm) & p95 (mm) & Outcome \\
\midrule
\textbf{a} Fully configured system ($n{=}22$)              & $\newm{0.62\pm0.12}$ & ${\sim}1.4$ & $\newm{22/22}$ complete \\
\textbf{b} Frozen $\Jhat$, post-excitation ($n{=}3$)    & 
1.0     & ${\sim}2.9$         & $2/3$ diverge, 1 degraded  \\
\textbf{c} Frozen $\Jhat$, ${\sim}12$\,s tracking ($n{=}2$) & $\newm{1.1}$ & ${\sim}\newm{3.6}$ & $1$ diverges, $1$ degraded \\
\textbf{d} Frozen $\Jhat$, ${\sim}30$\,s tracking ($n{=}4$) & $\newm{0.62\pm0.06}$ & ${\sim}1.4$ & $4/4$ complete \\
\textbf{e} No grip regularization ($n{=}10$) & $\newm{1.35}\pm\newm{1.13}$ & ${\sim}\newm{5.3}$ & $4$ diverge, $2$ degraded \\

\bottomrule
\end{tabularx}
\\[2pt]
{\scriptsize Single platform; observed values, not population statistics. Means, standard deviations, and p95 are reported over non-diverged runs only. }
\end{table}

\subsection{Sustained operation without recalibration}
\label{sec:sustained}
Because the Jacobian is re-estimated continuously, accuracy does not decay over
long runs. In a single uninterrupted ${\sim}31$-minute run ($26{,}600$
post-excitation steps) in which the robot wrote the entire alphabet, the in-plane error is stationary: across six equal time
windows the per-window mean stays within $0.59$--$0.69$\,mm and the 95th
percentile within $1.2$--$1.6$\,mm, with no upward trend; the rolling error
envelope, if anything, drifts slightly \emph{downward} as the estimator adapts
to slow grip changes (Fig.~\ref{fig:durability}, bottom). The
run could write the full alphabet without losing its grip, showing that a single grip sustains writing without
re-identification, demonstrating the robustness of an online estimate over one-shot
calibration.
Large excursions end in recovery rather than divergence: the
${\sim}20$\,mm spike during ``K'' (Fig.~\ref{fig:alphabet}) is absorbed
within seconds, without re-excitation or manual intervention.

\subsection{Evaluation on additional platforms in simulation}
\begin{figure}[htb]
  \centering
  \includegraphics[width=\linewidth]{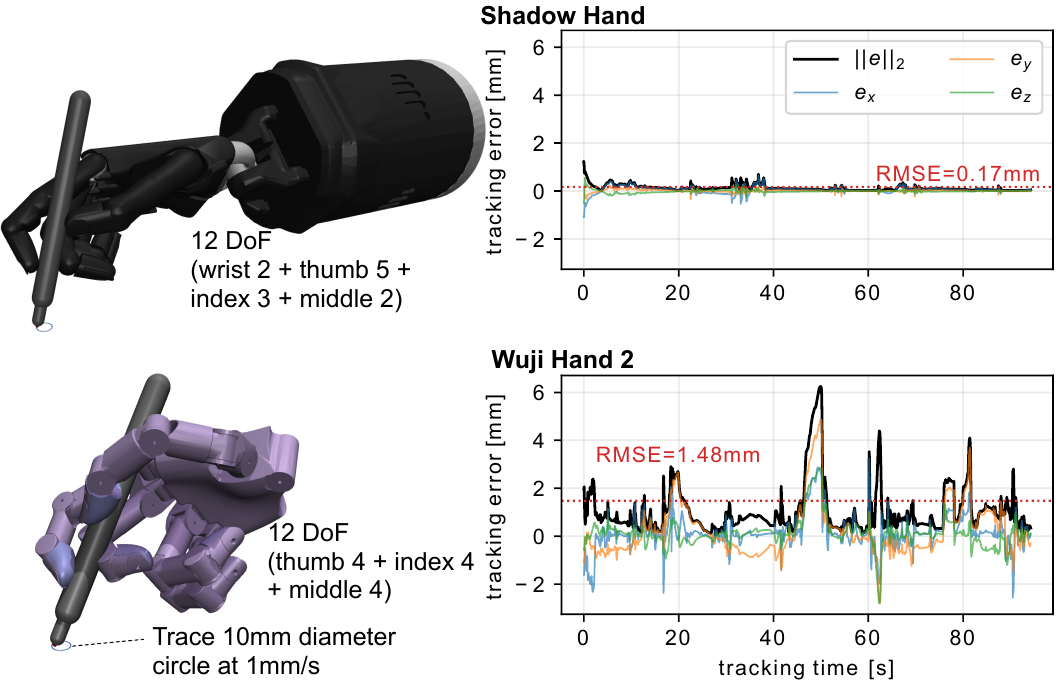}
  \caption{Overview and results of the trajectory tracking in simulation for two additional platforms. All other results in this paper were measured on a real robot platform.}
  \label{fig:sim_experiment}
\end{figure}

To evaluate the same estimator/controller formulation across hand embodiments, we ported it to the MuJoCo simulation environment and evaluated it on two additional robotic hands, the Shadow Hand and Wuji Hand 2. The simulation implementation is released open source in 
\href{https://github.com/srl-ethz/dexterity_from_jacobian}{this repository}.
Unlike the physical ORCA hand experiments, for which only the in-plane $(x,y)$ position is controlled, the simulation experiments controlled all three Cartesian coordinates $(x,y,z)$ of the pen tip ($m=3$). This was necessary because the simulated setup did not include the compliant TPU pen sleeve used on the physical system, making the grasp more susceptible to out-of-plane slippage. Thus, the simulation experiments evaluate a more difficult higher-dimensional tracking problem, while it is also important to note that they have access to noise-free ground-truth states.
An overview of the hand simulations and the results are shown in Fig.~\ref{fig:sim_experiment}. The Shadow Hand, which includes two wrist DoFs, had an RMSE of $0.17$\,mm, while the Wuji Hand 2 had a higher error of $1.48$\,mm.
These results provide initial evidence that the same online Jacobian-estimation and control approach can be applied to substantially different hand kinematics without requiring a hand-specific analytic Jacobian.

\section{Discussion}
\label{sec:discussion}

    Our result establishes that an online \emph{forward} task Jacobian, estimated live on hardware without a hand-object model, simulation training, or demonstrations, suffices to drive a contact-rich, continuous-trajectory in-hand writing task on a high-DoF anthropomorphic hand to achieve sub-millimeter in-plane tracking error in air and on paper. Its $0.64 \pm 0.10$ mm error is comparable in scale to the 0.41-0.45 mm free-space waypoint errors reported for a simpler symmetric three-fingered gripper~\cite{grace2024,grace2025}.

    We also compare our accuracy results to the closest \emph{RL-based} attempts for in-hand robotic writing. Zhao \emph{et al.}~\cite{Zhao2025-ar} and Hu \emph{et al.}~\cite{hu2023slender} each trained a three-fingered tactile hand in
    simulation, by deep reinforcement learning, to steer a grasped stick along
    trajectories (using only its fingers, as in our setup),
    reporting a stick-tip position error of $0.074$--$3.624$\,mm ~\cite{Zhao2025-ar} and $8.3$--$23.2$\,mm~\cite{hu2023slender}, respectively, with the former reporting the $x$ and $y$ errors separately. Both report errors only in the simulation environments they were trained in. Though some smaller and simpler trajectories produced lower tracking errors than ours, the errors were larger for more complex shapes, while our method resulted in errors of consistent magnitude for all trajectories tested.

    Further, the RL-based approaches demand heavy compute for training policies in simulation environments \cite{Zhao2025-ar, hu2023slender}.
    Our lightweight online estimate-based method evidently reaches, in the physical world, a precision regime that costlier RL-based approaches have so far shown only in simulation or, on hardware, without quantitative evaluation. The MuJoCo experiments also provide initial evidence that the formulation is not specific to the ORCA kinematics, as the same estimator/controller structure was applied to the Shadow Hand and Wuji Hand 2.
    
    Why an \emph{approximate} online estimate suffices is the key point. The ablations (Sec.~\ref{sec:ablations}) show that the online updating improves accuracy and robustness for long writing sessions, and approximate Jacobians are known to control well within a generous ``valid region'' of estimation error~\cite{grace2025,cheah2003}. Two design choices keep our estimate within it. The dedicated excitation phase
    gives the Jacobian an initial sense of which motor combinations move the pen across the task space (a difficulty unique to in-hand
    manipulation of an object, and absent from the serial-arm visual-servoing setting from which
    our update inherits~\cite{hosoda1994}), while continuous learning
    fine-tunes it throughout operation.

    The result is an in-plane steady-state map that is
    \emph{well}-conditioned ($\sigma_1/\sigma_2\approx2$ on average, below ${\sim}3.5$
    across runs), indicating that both in-plane directions were excited and subsequently learned similarly.

\textbf{Limitations.}
In the physical ORCA experiments,  the controller commands a 2-D action space parallel to the paper; the out-of-plane $z$ was kept uncontrolled, and on-paper writing relied on elevated paper to absorb the resulting ${\sim}2$--$3$\,mm $z$ drift. As a result, the current system cannot write on a rigid surface or perform multi-stroke writing (e.g.,\ ``i'') without arm assistance.
For the writing demo in Fig.~\ref{fig:teaser}, the robot arm was used to run a hardcoded motion to move the hand to the next character, but a full hand-arm coordinated system is out of scope for this work. Subsequent projects should explore how to distribute the control commands across the hand-arm system so that the fingers are used to achieve detailed motion while the arm provides gross positioning.

Also, we cannot handle contact-discontinuous tasks in which contact between the hand and object is broken and re-established, such as to (re)grasp a pen within the hand. The current formulation cannot plan and execute such discontinuous motions, as only the Jacobian of a continuous task with an unchanging contact state is being estimated. It is an open question how to extend such rapid online learning techniques to handle grasping or finger gaiting motions.

Writing is also slow ($\lVert \dot{\bm x}^\star \rVert_2=8\times10^{-4}$\,m/s): the speed ablation
(Sec.~\ref{sec:ablations}) shows pushing it faster trades robustness for
throughput (even a $2\times$ increase makes tracking unreliable), so reaching human-level writing speeds will require a controller that can compensate for the larger friction encountered there.
Accuracy is further bounded by persistent ArUco pen-marker jitter we could not remove; a different detector or a better-instrumented setup (higher frame rate, cleaner background) could improve it.

Finally, all reported errors are measured through the same vision pipeline that drives the controller; we did not independently verify the written output against a physical ground-truth (e.g., measuring the deposited ink), so the absolute accuracy values inherit the accuracy of the camera calibration and marker tracking.

\section{Conclusion}
\label{sec:conclusion}

We demonstrate that a rapid, embodied learning approach to dexterous manipulation can drive contact-rich in-hand writing on the anthropomorphic, tendon-driven ORCA hand by finger motion alone, after approximately 18\,s of initialization, while continuing to adapt during writing. Simulation experiments on two additional hand embodiments show the same formulation operating across substantially different hand kinematics.
This fast, adaptive, and embodied approach could achieve a level of robotic writing accuracy on biomimetic hands not yet seen in approaches such as RL or IL that require extensive modeling or data-collection efforts, suggesting that a lack of a precise model or data is not necessarily what prevents real dexterity.
Its speed and computational lightness further make it promising for applications where high-performance computing is unavailable, such as onboard prosthetics control.

\bibliographystyle{IEEEtran}
\bibliography{references}

\end{document}